\documentclass[a4paper, 10pt, conference]{ieeeconf}      % Use this line for a4
\usepackage{FG2024}
\usepackage{amssymb}
\usepackage{amsmath}
\usepackage{multirow}
\usepackage{verbatim}
\usepackage{enumerate}

\FGfinalcopy % *** Uncomment this line for the final submission

\IEEEoverridecommandlockouts                              % This command is only
\usepackage{graphics} % for pdf, bitmapped graphics files
\usepackage{epsfig} % for postscript graphics files
\usepackage{mathptmx} % assumes new font selection scheme installed
\usepackage{times} % assumes new font selection scheme installed

\def\FGPaperID{62} % *** Enter the FG2024 Paper ID here

\title{\LARGE \bf
DrFER: Learning Disentangled Representations for \\ 3D Facial Expression Recognition
}

\author{\parbox{16cm}{\centering
    {\large Hebeizi Li$^1$, Hongyu Yang$^{2,3}$, Di Huang$^1$}\\
    {\normalsize
    $^1$ School of Computer Science and Engineering, Beihang University, Beijing, China\\
    $^2$ Institute of Artificial Intelligence, Beihang University, Beijing, China\\
    $^3$ Shanghai Artificial Intelligence Laboratory, Shanghai, China}}
}

\usepackage{fancyhdr}
\begin{document}

\ifFGfinal
\thispagestyle{empty}
\pagestyle{empty}
\else
\author{Anonymous FG2024 submission\\ Paper ID \FGPaperID \\}
\pagestyle{plain}
\fi
\maketitle

\thispagestyle{fancy}

%%%%%%%%%%%%%%%%%%%%%%%%%%%%%%%%%%%%%%%%%%%%%%%%%%%%%%%%%%%%%%%%%%%%%%%%%%%%%%%%
\begin{abstract}

Facial Expression Recognition (FER) has consistently been a focal point in the field of facial analysis. In the context of existing methodologies for 3D FER or 2D+3D FER, the extraction of expression features often gets entangled with identity information, compromising the distinctiveness of these features. To tackle this challenge, we introduce the innovative DrFER method, which brings the concept of disentangled representation learning to the field of 3D FER.
DrFER employs a dual-branch framework to effectively disentangle expression information from identity information. Diverging from prior disentanglement endeavors in the 3D facial domain, we have carefully reconfigured both the loss functions and network structure to make the overall framework adaptable to point cloud data. This adaptation enhances the capability of the framework in recognizing facial expressions, even in cases involving varying head poses. Extensive evaluations conducted on the BU-3DFE and Bosphorus datasets substantiate that DrFER surpasses the performance of other 3D FER methods.
\end{abstract}

%%%%%%%%%%%%%%%%%%%%%%%%%%%%%%%%%%%%%%%%%%%%%%%%%%%%%%%%%%%%%%%%%%%%%%%%%%%%%%%%
\section{INTRODUCTION}

%Facial expression recognition (FER) is a crucial research topic in the field of face analysis, as it provides insights into human emotions, thoughts, and behaviours. 
%FER technique finds its applications in numerous real-world scenarios, such as traffic safety, psychological analysis, and human-computer interaction.
Facial Expression Recognition (FER) represents a crucial research area within the domain of facial analysis, playing a pivotal role in elucidating human emotions, cognitive processes, and behavioral patterns. FER techniques find applications in various real-world contexts, such as psychological analysis, human-computer interaction, and traffic safety, thereby underscoring its significance in both scientific research and practical applications.

The advancements in computer vision and image processing technologies have led to significant developments in FER. 
The origin can be traced back to the 1970s, when Ekman and Friesen~\cite{ekman1971constants} identified and defined six fundamental human facial expressions, including anger, disgust, fear, happiness, sadness, and surprise. 
Early FER approaches are mainly performed on 2D data, including static images \cite{shan2005robust}\cite{jabid2010local} and dynamic videos \cite{kotsia2006facial}\cite{zhao2007dynamic}, where both manual operator-based methods and machine learning solutions, such as deep neural networks, have been actively employed. 
However,  2D FER suffers from several intrinsic issues, \textit{i.e.,} illumination and viewpoint
variations, making its robustness problematic in the scenarios with high reliability requirements. 

With the advent of 3D data scanning equipment and rapid advancement of 3D modeling techniques, expression recognition using 3D data has emerged as a new trend in face analysis, enabling more thorough analysis of subtle facial movements. 
Moreover, since shape data are reputed to be insensitive to varying lighting and convenient for pose correction, it allows for the solutions of many issues that cannot be addressed in 2D FER.
%The key issue in 3D FER is the facial expression shape description, a description that is sufficiently discriminative but also excludes the interference of identity information.
In recent years, 3D FER methods have evolved into three main categories: model-based, feature-based, and deep learning-based approaches, which aim to enhance the representation of expression information from various perspectives, capitalizing on the advantages of 3D shape data.
Furthermore, there has been a rise in multi-modal FER methods that combine different clues from both 2D texture information and 3D geometry information for expression analysis. While significant progress has been made in these areas, there are still shortcomings in existing 3D FER and 2D+3D FER studies. Many of these methods primarily focus on enhancing the expressiveness of features related to geometric information or addressing domain gaps for multi-modal feature fusion. Unfortunately, they often mix expression features with identity information, resulting in a loss of distinctiveness in these features.

\begin{figure}[t]
\centering
\includegraphics[width=0.5\textwidth]{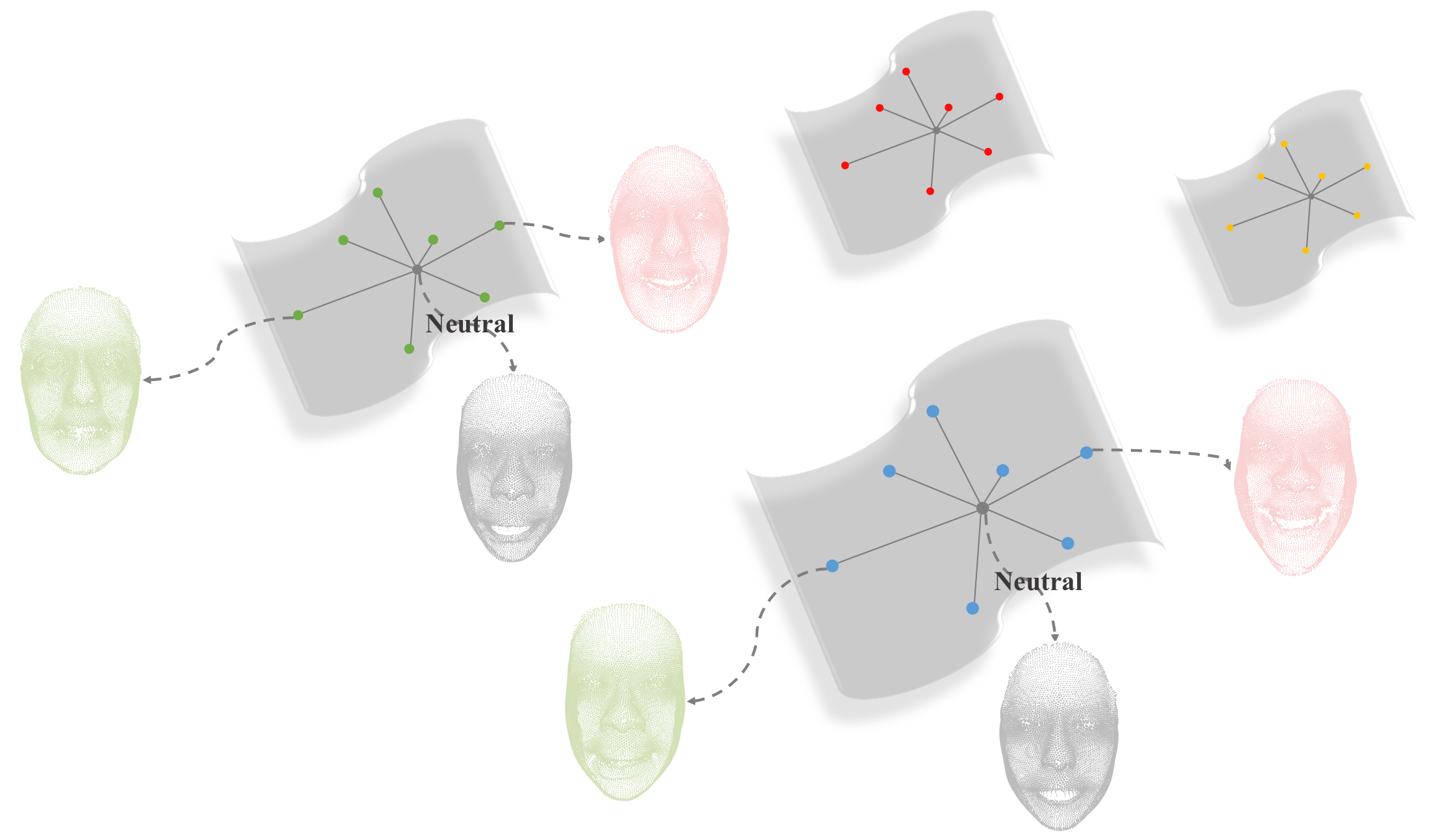}
\caption{Illustrations of 3D facial shape space. Human expressions occupy manifolds within a high-dimensional space, which exhibit similar patterns across different individuals and the center of each expression manifold corresponds to the neutral expression.}
\label{fig:facialspace}
\end{figure}

%Previous facial expression analysis \cite{chang2006manifold} stated that the latent variables representing identity and expression are lying on the manifold in high dimensional space. As illustrated in Figure \ref{fig:facialspace}, the expression manifolds are similar across individuals, making the goal of FER method to learn a universal expression representation. Moreover, Grasshof \textit{et al.} \cite{grasshof2017apathy}\cite{grasshof2020multilinear} found that apathy (the neutral expression in the datasets) is the center of all expressions, which means that the trajectories of obtaining various expressions by adapting the expression intensity all originate from this point.  This conclusion suggests that the implicit connection between different emotional faces with the same identity is the neutral face.  These discoveries provided two inspirations for the follow-up, one is that a universal expression representation can be constructed by learning the transformations from expressive to neutral faces, and the other is that identity information can be extracted by learning the neutral expressions of different people.
%While the disentanglement concept has found successful application in numerous 2D FER methods \cite{jiang2022disentangling}\cite{chang2021learning}\cite{liu2019hard} and has been effectively utilized in 3D face reconstruction tasks \cite{olivier2023facetunegan}\cite{jiang2019disentangled}, its integration into the 3D FER domain remains a novel development as of now.

Prior facial expression analysis \cite{chang2006manifold} has posited that the latent variables representing identity and expression exist on a manifold within a high-dimensional space. As depicted in Figure \ref{fig:facialspace}, the expression manifolds exhibit similarities across individuals, highlighting the goal of FER methods to learn a universal expression representation.
Furthermore, Grasshof \textit{et al.} \cite{grasshof2017apathy}\cite{grasshof2020multilinear} have observed that apathy (the neutral expression in the datasets) serves as the central point for all expressions, implying that the trajectories leading to various expressions, achieved by adjusting expression intensity, all originate from this neutral point. This discovery suggests an implicit connection between different emotional faces with the same identity, centered around the neutral face.
These findings provide two key inspirations for subsequent research. First, a universal expression representation can be constructed by learning the transformations from expressive to neutral faces. Second, the identity information can be extracted by learning the neutral expressions of different individuals. While such disentanglement concept has found successful applications in numerous 2D FER solutions \cite{jiang2022disentangling}\cite{chang2021learning}\cite{liu2019hard} and has been effectively utilized in 3D face reconstruction tasks \cite{olivier2023facetunegan}\cite{jiang2019disentangled}, its integration into the 3D FER domain represents a novel development as of now.

In this study, we present a novel method named DrFER, which leverages the observations above to disentangle facial expression representations from identity-related factors for 3D FER. In particular, unlike prior 3D disentanglement methods that primarily use mesh as source data, a key distinction of our approach is the use of facial point cloud data as input, chosen for its inherent robustness to variations in facial pose. To accommodate this choice, we have made substantial adaptations to both the configuration of the loss functions and the network architecture.
Specifically, to address the challenge that the features extracted from point clouds do not conform to the standard distribution, we choose not to use KL Loss and JS Loss, which are commonly employed as disentanglement supervisions. Instead, we carefully design the loss functions to regulate the latent space. Additionally, the network architecture, based on PointNet++ \cite{qi2017pointnet++}, is thoughtfully designed to achieve effective disentanglement and re-coupling.
These adaptations enable the effective application of the disentanglement framework to point cloud data, enhancing its robustness and performance in 3D FER.
%\begin{comment}
%Two branches are configured to extract expression information and identity information respectively, which are capable of reconstructing the mean expression face and the neutral face. 
%When they are connected in a crosswise manner, both branches will eventually reconstruct the mean neutral face; when they are connected together with a fusion module, the whole pipeline will reconstruct the original face.
%With the multi-stage training strategy and the well-designed constraints, our method disentangles the expression information from the identity information and alleviates the influence of irrelevant factors on expression recognition.
%\end{comment}
The contributions of this study can be briefly summarized as follows:
\begin{itemize}
    \item   
   This paper introduces an innovative approach, DrFER, which marks the first application of the disentanglement paradigm within the field of 3D FER. 
    \item 
   The framework represents a notable advancement over prior endeavors, primarily through the loss functions and network architecture adeptly designed for the analysis of 3D point cloud data.   
    \item The proposed approach achieves a state-of-the-art level of performance on both the BU-3DFE and Bosphorus datasets, positioning it alongside the outcomes achieved by other 2D+3D FER techniques. Importantly, DrFER demonstrates robustness in handling rotational poses. 
\end{itemize}

%\begin{table}
%\caption{An Example of a Table}
%\label{table_example}
%\begin{center}
%\begin{tabular}{|c||c|}
%\hline
%One & Two\\
%\hline
%Three & Four\\
%\hline
%\end{tabular}
%\end{center}
%\end{table}

%%%%%%%%%%%%%%%%%%%%%%%%%%%%%%%%%%%%%%%%%%%%%%%%%%%%%%%%%%%%%%%%%%%%%%%%%%%%%%%%
\section{RELATED WORK}

\subsection{3D Facial Expression Recognition}

Wang \textit{et al.} \cite{wang20063d} in their pioneering work laid the foundation for utilizing 3D data in facial expression recognition. Subsequently, a number of methods have been developed with a focus on extracting 3D facial geometric features for FER and they can be broadly categorized into two major types: model-based \cite{mpiperis2008bilinear}\cite{gong2009automatic} and feature-based \cite{yang2015automatic}\cite{tang20083d}\cite{zhen2016muscular}. Model-based methods typically derive expression information by modeling generalized shape statistics and fitting 3D facial expression samples to retrieve the corresponding fitting coefficients. For instance, Mpiperis  \textit{et al.} \cite{mpiperis2008bilinear} introduced a bilinear elastically deformable model capable of estimating both identity and expression parameters, with the latter being utilized for expression recognition. Gong \textit{et al.} \cite{gong2009automatic} partitioned the 3D face into a Base Face Shape Component and an Expression Shape Component, where the expression-related component was employed to extract expression features. On the other hand, feature-based approaches applied various feature operators on 3D scans or computed corresponding 2D shape attribute maps. These numerical features are then utilized by a classifier to predict the FER results. For example, Yang \textit{et al.} 
\cite{yang2015automatic} computed scattering features on depth, normal, and shape index maps and used SVM for expression classification. However, the FER accuracy of  these methods heavily rely on the design of these operators or hand-crafted features.

\begin{figure*}[htb]
\centering
\includegraphics[width=1.01\textwidth]{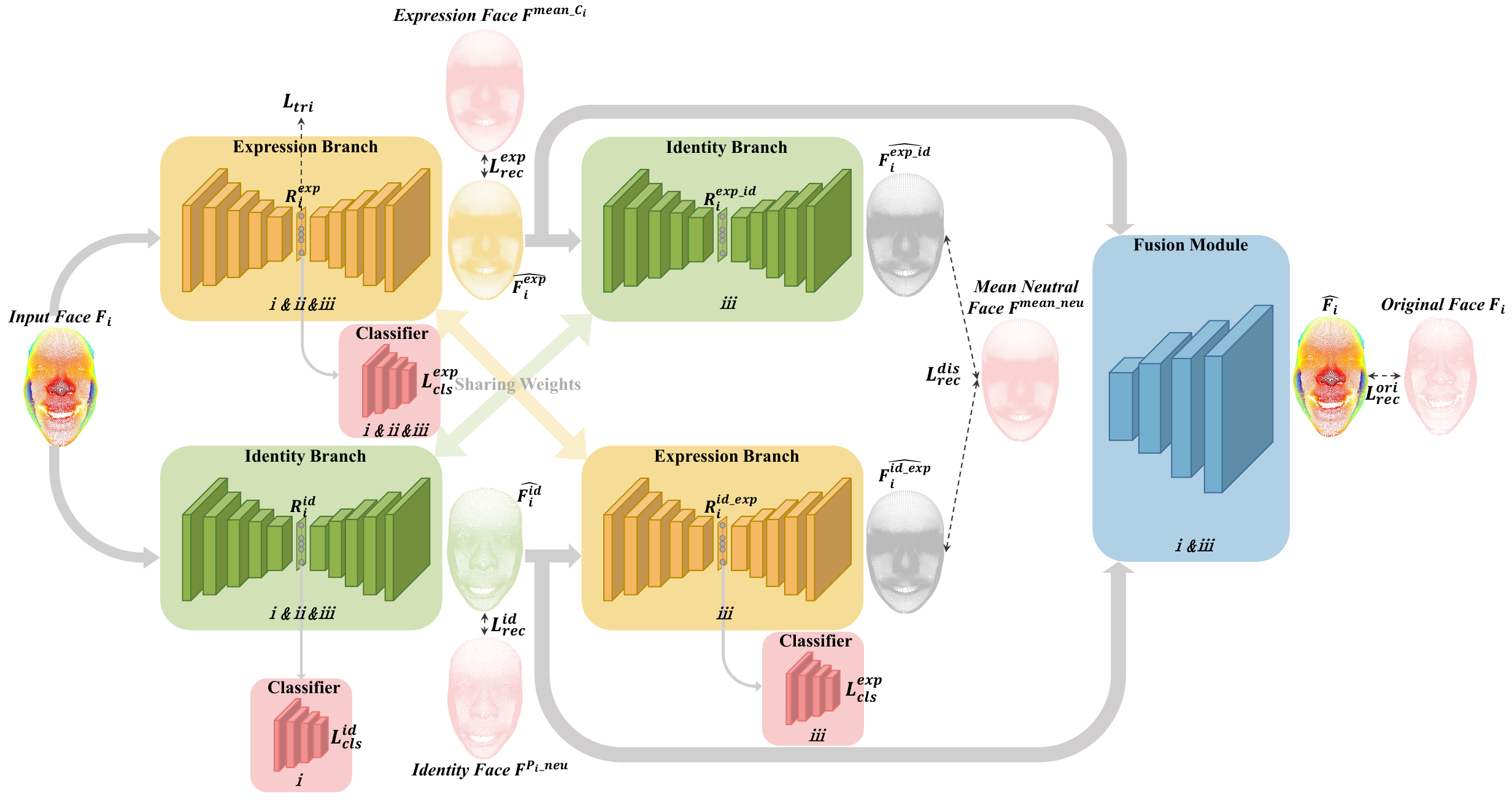}
\caption{Method overview. The proposed DrFER model comprises two key components: the disentangling component and the fusion component. The former employs a dual-branch architecture to explicitly learn expression and identity features, and generate the corresponding de-identity and de-expression faces, respectively. The model subsequently recombines these disentangled faces in a cross-over manner and reconstruct the original face with the fusion component, facilitating the disentanglement process. To guide the training process effectively, a series of training losses are employed, including those specifically tailored for point cloud data. The training stages corresponding to each module are labeled with lowercase Roman numerals in the figure.}
\label{fig:framework}
\end{figure*}

With the rapid advancements in deep learning technology, its applications have extended into the filed of 3D facial expression recognition. Notably, Chen \textit{et al.} \cite{chen2018fast} presented an innovative  model known as the Fast and Light Manifold CNN (FLM-CNN). Their approach involved a modification of the weighting mechanism, inspired by human vision, resulting in the development of a highly precise and rotation-invariant model for FER.
While there exist numerous deep learning methods that leverage 2D+3D multimodal data, the wealth of geometric information inherent in pure 3D data remains an unexplored frontier with significant potential for further investigation.

%In summary, Wang et al.'s pioneering work paved the way for 3D facial expression recognition, with subsequent methods falling into the categories of model-based and feature-based approaches. More recently, the integration of deep learning, as exemplified by the FLM-CNN model by Chen et al., has brought about significant advancements in this field.

\subsection{Multi-modal Facial Expression Recognition}

To combine 2D texture information and 3D geometric information, several studies have incorporated 2D and 3D multi-modal data to enhance the accuracy of FER. In particular, Li \textit{et al.} \cite{li2015efficient} combined local texture descriptors and shape operators with multistage gradients, employing SVM for training and classification. In another work, Li  \textit{et al.} \cite{li2017multimodal} pioneered the use of Convolutional Neural Networks (CNN) in 2D + 3D FER. Their proposed DF-CNN extracted features from various sources, including texture maps, depth maps, normal vector maps, and shape index maps, which were then stacked and fed into the network for expression classification. 
Zhu \textit{et al.} \cite{zhu2019discriminative} introduced a DA-CNN model incorporating an attention mechanism to enhance classification ability of the network, while Jiao \textit{et al.} \cite{jiao2019facial} presented FA-CNN, which leveraged attention mechanism to identify discriminative facial regions within multi-modal data.
%In \cite{jiao20202d+}, the researchers  incorporated a novel map generation method from the viewpoint of information theory, and slight expression variations from the discriminative dynamic range are comprehensively enhanced by maximizing the entropy under the depth distortion constraint.
%Then local facial parts can be efficiently generated for multi-scale feature learning. 
Besides,  Lin \textit{et al.} \cite{lin2020orthogonalization} proposed OGF$^{2}$Net, which employed orthogonal loss to guide the fusion of multi-modal features.
Sui \textit{et al.} conducted a series of studies \cite{sui2021ffnet}\cite{sui2023afnet} where they utilized facial landmarks as priors for generating masks, strengthening the capacity to extract features. In their latest work \cite{zhu2022cmanet}, they improved the mask to a curvature-aware soft mask and integrated a multi-modal attention fusion module, achieving state-of-the-art performance.

%To summarize, the workflow of current multi-modal FER methods typically consists of a feature extraction stage employing distinct networks for 2D and 3D data, succeeded by feature fusion facilitated by an additional module. However, prior research has primarily emphasized the improvement of representation within each modality and the fusion of features across modalities. It is important to note that directly extracted expression features often suffer from contamination due to interference from identity-related features.
%In current multi-modal FER methods, the workflow usually involves separate networks for extracting features from 2D and 3D data, followed by feature fusion using an additional module. However, previous research has mainly focused on enhancing representation within each modality and fusing features across modalities. It is worth noting that directly extracted expression features are often contaminated by interference from identity-related features.
In current multi-modal FER methods, the workflow typically includes utilizing distinct networks to extract features from 2D and 3D data, followed by feature fusion through an additional module. However, the existing research has primarily emphasized improving representation within each modality and fusing features across modalities, without explicitly separating the identity information mixed in the extracted expression features.

\subsection{Disentangled Face Representations}

Recently, there has been a growing interest in applying disentanglement techniques to face-related applications, including age-invariant face recognition \cite{wang2019decorrelated}\cite{huang2021age}, face editing \cite{tran2017disentangled}\cite{zheng2021unsupervised}\cite{deng2020disentangled}, and face reenactment \cite{huang2020learning}. 
Particularly, in the 3D domain, a number of studies have delved into 3D face reconstruction \cite{olivier2023facetunegan}\cite{jiang2019disentangled} with a focus on achieving precise results through the disentanglement of expression and identity variables. These investigations underscore the effectiveness of the disentanglement concept in the field of 3D face analysis.

Furthermore, there are several studies exploring FER via disentangled representation learning \cite{jiang2022disentangling}\cite{chang2021learning}\cite{liu2019hard}. For instance, Jiang  \textit{et al.}  \cite{jiang2022disentangling} proposed a method that decomposes pose, identity, and expression information in the embedding space, leading to improved FER accuracy. Chang \textit{et al.} \cite{chang2021learning} augmented their disentangled network with optical flow information and enforced constraints on the reconstruction results through cycle consistency loss. Liu \textit{et al.}  \cite{liu2019hard} adopted a metric learning approach to extract expression information by comparing original expression images with the generated neutral face images. These studies convincingly demonstrated the potential benefits of disentanglement techniques in enhancing expression recognition accuracy.
%To the best of our knowledge, this paper marks the first attempt to present a disentangled representation learning method for 3D FER, where our research efforts have been dedicated to adapting the disentanglement framework to effectively work with 3D facial point cloud data.
To the best of our knowledge, this paper represents the pioneering effort to introduce a disentangled representation learning approach specifically designed for 3D FER. 

%\addtolength{\textheight}{-3cm}   % This command serves to balance the column lengths

%%%%%%%%%%%%%%%%%%%%%%%%%%%%%%%%%%%%%%%%%%%%%%%%%%%%%%%%%%%%%%%%%%%%%%%%%%%%%%%%

\section{METHOD}

%This section describes in detail the proposed approach for 3D FER by learning disentangled representation. We introduce below the overview of entire framework, the network architecture, the training strategy and the loss funtion, respectively.

%This section provides an in-depth description of the proposed DrFER method which achieves accurate and robust 3D expression recognition by learning disentangled representations. Below, we outline the entire framework, elaborate on the network architecture, delve into the training strategy, and discuss the loss function utilized, respectively.
This section provides a detailed description of the proposed method. We outline the framework, network architecture, training strategy, and loss function used.

\subsection{Overview}

The proposed disentanglement approach comprises two key components, \textit{i.e.}, the disentangling component and the fusion component, as illustrated in Figure \ref{fig:framework}. Given that diverse expressions originate from a neutral face of an individual, the disentangling component 
 of DrFER thus employs a dual-branch network with two identical encoders to explicitly learn expression and identity features, respectively. The model follows the general disentanglement paradigm, intertwining the produced de-identity and de-expression faces in a cross-over fashion and re-generate 3D faces with the fusion component. These operations not only facilitate the disentanglement learning process but also ensure the comprehensive capture of essential information. To guide the training process towards enhanced accuracy, a range of training losses is employed, including the customized Chamfer distance metric and triplet loss tailored for 3D point cloud data. Upon model convergence, the extracted expression feature is fed into a classification head to make the final decision.

\subsection{Network Architecture}
\begin{figure}[htb]
\centering
\includegraphics[width=0.5\textwidth]{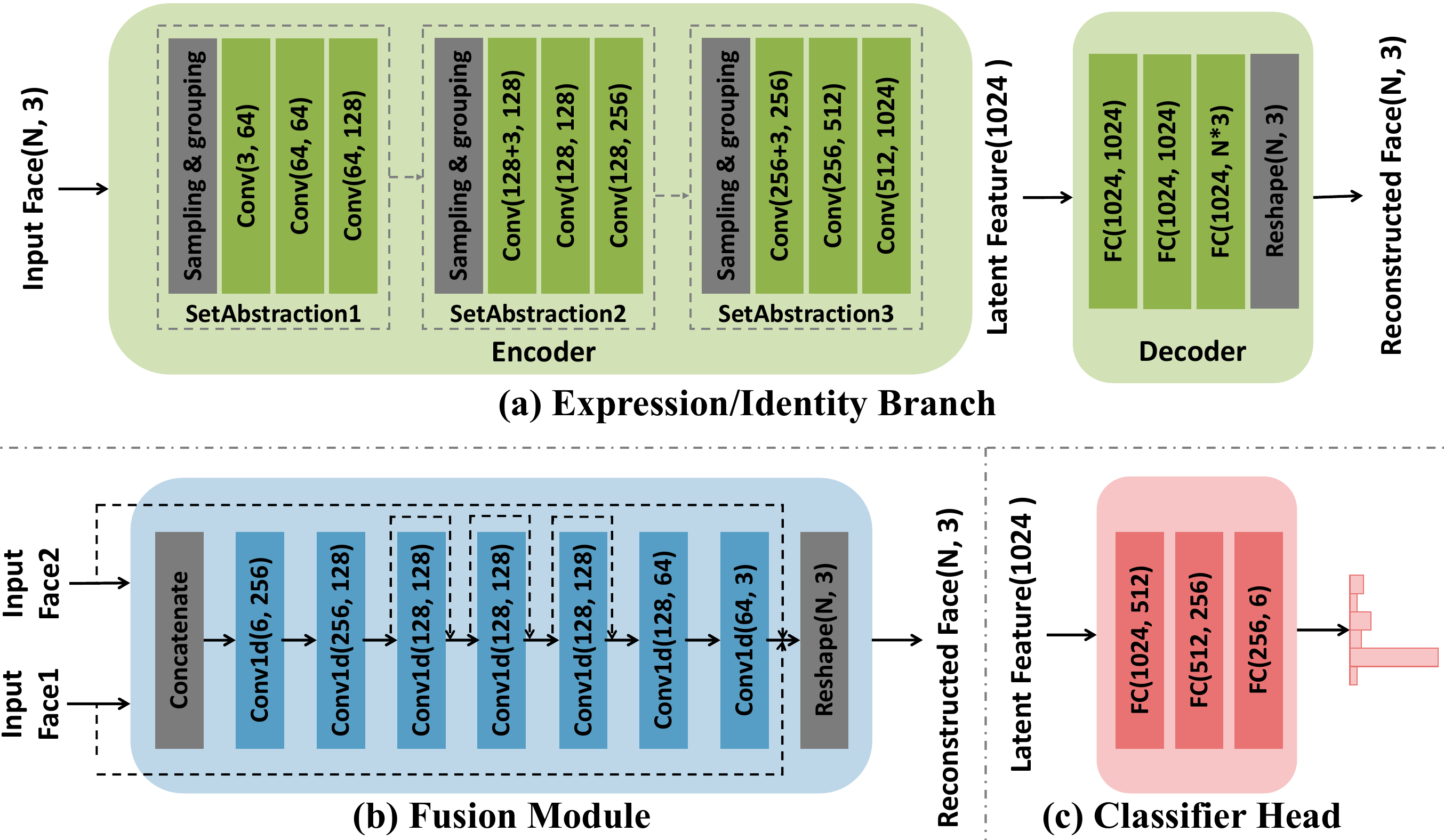}
\caption{Detailed architectures of the proposed expression/identity branch, the cross-modal fusion module, and the classifier.}
\label{fig:network}
\end{figure}

Formally, the input 3D face point cloud is represented as $F_{i}\in \mathbb{R}^{n\times3}\left(i\in\left[1,\dots,m\right]\right)$, where $n$ represents the number of points, and $m$ is the count of input 3D face scans. Since the model operates under the assumption that expressions and identities inhabit distinct and mutually independent feature spaces, two dedicated branches, $\Phi_{exp}$ and $\Phi_{id}$, are employed to handle expression and identity information in  $F_{i}$, respectively. 
Both branches share an identical \textit{Encoder-Decoder} structure as illustrated in Figure \ref{fig:network} (a), which is implemented based on Pointnet++ \cite{qi2017pointnet++}.
The encoder contains three \textit{set abstraction} levels, and each is made of three key layers: \textit{Sampling layer}, \textit{Grouping layer} and \textit{Pointnet layer}. 
The \textit{Sampling layer} selects a set of centroid points from the input points, \textit{Grouping layer} searches for neighboring points of the centroids, and \textit{Pointnet layer} encodes the local region patterns into feature vectors.
The decoder encompasses multiple fully-connected layers that decode the high-dimensional latent features into a 3D face point cloud. To summarize, the two branches produce latent representations, namely $R_{i}^{exp}$ and $R_{i}^{id} \in \mathbb{R}^{n\times d}$ (with $d$ representing the dimension) in the feature space, as well as the corresponding faces $\hat{F_{i}^{exp}}$ and $\hat{F_{i}^{id}}$ in the observation space. Notably, the expression feature is denoted as residing in a 1024-dimensional space.

To ensure that the two branches learn independent information, they are interconnected in a cross-over manner, leading to the derivation of new latent representations $R_{i}^{exp\_id}$ and $R_{i}^{id\_exp}$, along with newly reconstructed faces $\hat{F_{i}^{exp\_id}}$ and $\hat{F_{i}^{id\_exp}}$. For instance, in the upper stream of Figure \ref{fig:framework}, input $F_i$ is fed into the expression branch $\Phi_{exp}$ to yield a face $\hat{F_{i}^{exp}}$ containing only the original expression information. Subsequently, $\hat{F_{i}^{exp}}$ is passed through the identity branch $\Phi_{id}$ to generate a face $\hat{F_{i}^{exp\_id}}$ devoid of both expression and identity information. 
%The reconstructed $\hat{F_{i}^{id\_exp}}$ from the lower stream should be similar.
Ideally, after the operations above, the representations of both streams do not retain information specific to the given input face and the mean neutral face $F^{mean\_neu}$ should be precisely reconstructed. Notably, no supplementary modules are introduced during the cross-connection process, it is achieved by sharing weights with the original dual-branch architecture, thereby facilitating their concurrent optimization.

%Given that both branches exhibit \textit{encoder-decoder} architectures, and their inputs and outputs consist of face point clouds, direct connections can be established between them without necessitating the introduction of  when executing the cross-over fashion.

Moreover, the outputs of the two original disentanglement branches can be fused through a fusion module, denoted as $\Upsilon$, to reconstruct the original input face, represented as $\hat{F_i}$. The architecture of this fusion module is depicted in Figure \ref{fig:network} (b), which can generate a face corresponding to the expression information from $Face\_1$ and the identity property from $Face\_2$. The primary objective of the fusion module is to assimilate the two input face point clouds by harmonizing their respective feature components, ultimately reconstructing a comprehensive facial representation. 
When the face pair consists of faces $\hat{F_{i}^{exp}} $ and $\hat{F_{i}^{id}}$,  $F_i$ should accurately reconstructed. 
In particular, to strike a balance between the high-level semantic information and the observed point cloud information,  skip connections have been incorporated at multiple scales within the fully connected layers of the fusion module. 
This strategic inclusion ensures that the module captures an ample amount of information during feature extraction.

Given that our primary objective is facial expression recognition utilizing the expression feature $R_{i}^{exp}$, we include an auxiliary classifier within the expression branch to oversee and guide the learning process. The architecture of this classifier is depicted in Figure \ref{fig:network} (c).
%Considering that our ultimate goal is to perform FER using the expression feature $R_{i}^{exp}$, an auxiliary classifier is incorporated into the expression branch to supervise the learning process. The classifier architecture is displayed in Figure \ref{fig:network} (c).
%\textbf{A classifier with a similar structure is employed in pretraining the identity branch.}

%In order to emphasize the network has learned the complete facial information, we use the fusion module $\Upsilon$ to fuse the reconstructed expression face $\hat{F_{i}^{exp}}$ and identity face $\hat{F_{i}^{id}}$ to construct the original input face $F_i$.
%Ultimately, the disentangled expression feature extracted from the well-trained expression branch, $R_{i}^{exp}$, is harnessed for 3D FER.

\subsection{Training Strategy \& Loss Functions}

We apply a multi-stage strategy to stablize the training process of the whole network, including the following steps:
\begin{enumerate}[i.]
    \item Pretrain the encoders of $\Phi_{exp}$ and $\Phi_{id}$ with two specific classifiers and pretrain the fusion module $\Upsilon$ with the ground-truth face scans as supervision.
    \item Fine-tune the two branches $\Phi_{exp}$ and $\Phi_{id}$ individually using the reconstruction constraints.
    \item Perform end-to-end training with a very small learning rate after joining the cross-over setup and the fusion module $\Upsilon$.
\end{enumerate}

To be specific, considering that pre-training in categorization is the most direct way for the encoders to learn discriminative features, we introduce the expression and identity recognition tasks to pre-train the encoders in the first phase.
Let $\varepsilon_{exp}$ and $\varepsilon_{id}$ be the auxiliary classifiers for the expression and identity branches, respectively, the loss terms are defined as:
\begin{align}
    L_{stage\romannumeral1}^{exp} &= L_{cls}^{exp} = \ell _{CE}(\varepsilon _{exp}(R_i^{exp}), C_i), \\
    \label{eq_ce}
    L_{stage\romannumeral1}^{id} &= L_{cls}^{id} = \ell _{CE}(\varepsilon _{id}(R_i^{id}), P_i),
\end{align}
where $\ell _{CE}$ denotes the cross-entropy loss and $C_i$ and $P_i$ indicate the corresponding ground-truth labels of the sample $F_i$.
Furthermore, the fusion module $\Upsilon$ rebuilds the original face using expression-identity face pairs. 
In particular, we group the faces by labels $C_i$,  $P_i$ and calculate the mean faces $F^{mean\_C_i}$ and $F^{P_i\_neu}$.  The pair ($F^{mean\_C_i}, F^{P_i\_neu}$) is used as the input to the fusion module to reconstruct $F_i$. 
The reconstruction loss is formulated as:
\begin{equation} \label{eq_rec}
    L_{stage\romannumeral1}^{fusion} = L_{rec}^{ori} = d_{CD}(\hat{F_i}, F_i),
\end{equation}
where $d_{CD}$ represents the chamfer distance \cite{fan2017point} between two point clouds.

In the second stage, we leverage the reconstruction task to fine-tune the encoders and train their respective decoders. This strategy aligns with insights from previous studies \cite{grasshof2017apathy,grasshof2020multilinear}, which posit that the neutral face of an individual serves as the central point within the corresponding high-dimensional expression space. This neutral face can be regarded as predominantly containing identity information, thus making it a suitable representation of identity alone.
Conversely, from a statistical analysis perspective, the average face corresponding to a specific expression type is assumed to lack identity information and predominantly encapsulate expression-related details.
Hence, we utilize $F^{mean\_C_i}$ and $F^{P_i\_neu}$ as ground-truth representations 
for the disentanglement of these two kinds of information through a reconstruction task. This training loss can be defined as follows:

\begin{align}
    L_{rec}^{exp} &= d_{CD}(\hat{F_i^{exp}}, F^{mean\_C_i}), \\
    L_{rec}^{id} &= d_{CD}(\hat{F_i^{id}}, F^{P_i\_neu}).
\end{align}
To further constrain the feature space learned by the expression branch, we exploit a triplet loss \cite{schroff2015facenet} which brings samples of the same class closer together and samples of different classes farther apart. This  triplet loss is represented as:
\begin{equation} \label{eq1}
    L_{tri} = \max \left ( \left \| R_A-R_P \right \|^2 -\left \| R_A-R_N \right \|^2+\alpha , 0 \right ),
\end{equation}
where $R_A$ is the feature of an anchor sample $A$, $R_P$ and $R_N$ denotes the features of the positive and negative samples, respectively, and $\alpha$ is a hyper-parameter to bound the distance between the positive and negative classes.
It is important to highlight that our approach employs the triplet loss and classification loss instead of the commonly used KL divergence loss in prior 3D disentanglement frameworks designed for mesh data as input. The reason behind this choice lies in the fact that, unlike the explicit expression and identity variables present in 3D Morphable Models (3DMM), it is not straightforward to apply Gaussian distribution regularity (\textit{N}(0,1)) to the expression features extracted from the point cloud data. This consideration motivates the adoption of the triplet loss, a commonly used metric in face recognition, to effectively disentangle the features.
Overall, the losses in the second stage include:
\begin{align}
    L_{stage\romannumeral2}^{exp} &= L_{rec}^{exp} + L_{tri} + L_{cls}^{exp}, \\
    L_{stage\romannumeral2}^{id} &= L_{rec}^{id}.
\end{align}

In the third stage, we further impose constraints on the outcomes of the cross-over and fusion reconstructions. In cases where the cross-over structure is utilized, both the upper and lower streams are compared against the ground-truth $F^{mean\_neu}$. To assess the accuracy of reconstruction, we employ the Chamfer distance, which was introduced in the second stage, and denote it as:
\begin{equation}
    L_{rec}^{dis} = d_{CD}(\hat{F_i^{exp\_id}}, F^{mean\_neu}) + d_{CD}(\hat{F_i^{id\_exp}}, F^{mean\_neu}).
\end{equation}
Besides, the reconstruction loss of the fusion module is consistent with \eqref{eq_rec}.
In summary, the comprehensive loss function employed in the third stage of the end-to-end training is formulated as:
\begin{equation} \label{eq2}
    L_{stage\romannumeral3} = \underbrace{L_{cls}^{exp} + L_{tri} + L_{rec}^{exp}}_{\Phi_{exp}} + \underbrace{L_{rec}^{id}}_{\Phi_{id}} + \underbrace{L_{rec}^{dis}}_{\Phi_{exp},\Phi_{id}} + \underbrace{\lambda L_{rec}^{ori}}_{\Phi_{exp},\Phi_{id},\Upsilon},
\end{equation}
where $\lambda$ is a hyper-parameter used for balancing the training critics.
The various network modules that are supervised by the losses are labeled below the equation.

%   \begin{figure}[thpb]
%      \centering
%      %\includegraphics[scale=1.0]{figurefile}
%      \caption{Inductance of oscillation winding on amorphous
%       magnetic core versus DC bias magnetic field}
%      \label{figurelabel}
%   \end{figure}

%%%%%%%%%%%%%%%%%%%%%%%%%%%%%%%%%%%%%%%%%%%%%%%%%%%%%%%%%%%%%%%%%%%%%%%%%%%%%%%%

\section{EXPERIMENTS}

In this section, we present the results of the proposed DrFER model on the publicly available BU-3DFE and Bosphorus datasets. We aim to validate the effectiveness of our method through a comparison with other state-of-the-art approaches and extensive ablation studies. The following sections provide details on the datasets used, implementation details, experimental results, and visualizations of the outcomes.

\subsection{Datasets}

\textit{BU-3DFE \cite{yin20063d}}: 
BU-3DFE is the most commonly used publicly available dataset for academic research on 3D FER, which contains 2,500 scans of 100 subjects between the ages of 18 and 70 (56 females and 44 males).
Each subject has 25 samples of seven expressions in total: one neutral sample and six basic expressions samples of four different intensities.
We follow a consistent experimental protocol with previous work \cite{chen2018fast}\cite{zhu2022cmanet}\cite{sui2023afnet}\cite{lin2020orthogonalization} for fair comparisons.
In this protocol, we randomly select 60 subjects out of total 100 subjects and fix them during the entire experiment.
10-fold cross-validation experiments are conducted on the sub-dataset which consists of six expressions of the two highest intensities from these 60 subjects.
The cross-validation groups the 60 subjects into 10 subsets, each of which is selected once as the test set and the remaining 9 subsets are used as the training set.
This validation process is repeated for 100 times to minimize jitters in the experimental results.

\textit{Bosphorus \cite{savran2008bosphorus}}: 
%Bosphorus consists of 4,666 3D face scans from 105 subjects with ages ranging from 25 to 35. 
%There are 65 subjects containing the complete set of six basic expressions with one intensity, and we select 60 subjects from them for 10-fold cross-validation with reference to the protocol of BU-3DFE.
Bosphorust comprises a total of 4,666 3D face scans collected from 105 subjects within the age range of 25 to 35. Among these subjects, 65 individuals exhibit the complete set of six basic expressions, each with a single intensity level. For our experimental setup, we opt to utilize a subset of 60 subjects from this group, following the 10-fold cross-validation protocol, which is aligned with the approach used in the BU-3DFE dataset.

\subsection{Implementation Details}

%Regarding data preprocessing, we carry out rigid registration for all point clouds, aligning them with the same 3D mesh. Subsequently, we resample these point clouds to a total of 39,923 points.
%For each face scan, we select 2,048 points as network input, a choice aligned with the farthest point sampling (FPS) method utilized in Pointnet++\cite{qi2017pointnet++}. Additionally, we augment the data by applying random dropout and random scale transformations in each epoch.
Regarding data preprocessing, we perform rigid registration on all point clouds, aligning them to a common 3D mesh. Following this, we resample the point clouds to a uniform size of 39,923 points. For each face scan, we choose 2,048 points as input for the network, in line with the farthest point sampling technique as employed in PointNet++ \cite{qi2017pointnet++}. Furthermore, we augment the dataset by introducing random dropout and random scale transformations during each training epoch.

The network is trained using the Adam optimizer with beta values (0.9, 0.999). In Stage \romannumeral1, the initial learning rate is set to 0.001, and both branches and the fusion module are trained with a batch size of 24. In Stage \romannumeral2, the learning rate is adjusted to 0.0001 for fine-tuning the two branches with the reconstruction tasks. In Stage \romannumeral3, to mitigate the risk of overfitting, the learning rate is further reduced to 0.00001, and fine-tuning is performed on the entire network. During the fine-tuning phase, the cross-over structure is introduced, necessitating an adaptation of the batch size to 16.
The hyperparameters in \eqref{eq1} and \eqref{eq2} are set to $\alpha = 0.3$ and $\lambda = 0.1$.
Training is carried out within the PyTorch framework, and the model is validated using an NVIDIA GeForce RTX 3080 Ti GPU.

\begin{table}[htb]
\centering
\caption{Quantitative comparison on BU-3DFE. \dag indicates that the results achieved by using only 3D modality are also reported.}
\label{table:BU3D}
\begin{tabular}{cccc}
\hline
Method                        & Modality & Feature       & Accuracy(\%) \\ \hline
Li \textit{et al.} \cite{li2015efficient} & 2D+3D    & Hand-crafted  & 86.32    \\
Li \textit{et al.} \cite{li2017multimodal} & 2D+3D    & Deep Learning & 86.86    \\
Wei \textit{et al.} \dag \cite{wei2018unsupervised} & 2D+3D    & Deep Learning & 88.03    \\
Zhu \textit{et al.} \dag \cite{zhu2019discriminative} & 2D+3D    & Deep Learning & 88.35    \\
Jan \textit{et al.} \dag \cite{jan2018accurate} & 2D+3D    & Deep Learning & 88.54    \\
Zhu \textit{et al.} \dag \cite{zhu2020intensity} & 2D+3D    & Deep Learning & 88.75    \\
Ni \textit{et al.} \dag \cite{ni2022facial} & 2D+3D    & Deep Learning & 88.91    \\
Lin \textit{et al.} \dag \cite{lin2020orthogonalization} & 2D+3D    & Deep Learning & 89.05    \\
Jiao  \textit{et al.} \cite{jiao2019facial} & 2D+3D    & Deep Learning & 89.11    \\
Oyebade  \textit{et al.} \dag \cite{oyedotun2017facial} & 2D+3D      & Deep Learning & 89.31     \\ 
Jiao \textit{et al.} \cite{jiao20202d+} & 2D+3D    & Deep Learning & 89.72    \\
Sui \textit{et al.} \dag \cite{sui2021ffnet} & 2D+3D    & Deep Learning & 89.82    \\
Sui \textit{et al.} \dag \cite{sui2023afnet} & 2D+3D    & Deep Learning & 90.08    \\
Zhu \textit{et al.} \dag \cite{zhu2022cmanet} & 2D+3D    & Deep Learning & 90.24    \\ \hline
Wei \textit{et al.} \cite{wei2018unsupervised} & 3D       & Deep Learning & 74.44    \\
Tang \textit{et al.} \cite{tang20083d} & 3D       & Hand-crafted  & 74.51    \\
Gong \textit{et al.} \cite{gong2009automatic} & 3D       & Hand-crafted  & 76.22    \\
Ni \textit{et al.} \cite{ni2022facial} & 3D       & Deep Learning & 80.11    \\
Li \textit{et al.} \cite{li20123d} & 3D       & Hand-crafted  & 80.14    \\
Jan \textit{et al.} \cite{jan2018accurate} & 3D       & Deep Learning & 81.83    \\
Zhu \textit{et al.} \cite{zhu2022cmanet} & 3D       & Deep Learning & 84.03    \\
Zhen \textit{et al.} \cite{zhen2016muscular} & 3D       & Hand-crafted  & 84.50     \\
Oyebade \textit{et al.} \cite{oyedotun2017facial} 
& 3D & Deep Learning & 84.72    \\
Yang \textit{et al.} \cite{yang2015automatic} & 3D       & Hand-crafted  & 84.80     \\
Lin \textit{et al.} \cite{lin2020orthogonalization} & 3D       & Deep Learning & 85.20     \\
Chen \textit{et al.} \cite{chen2018fast} & 3D       & Deep Learning & 86.67    \\
Sui \textit{et al.} \cite{sui2023afnet} & 3D       & Deep Learning & 86.97    \\
Zhu \textit{et al.} \cite{zhu2020intensity} & 3D       & Deep Learning & 87.19    \\
Sui \textit{et al.} \cite{sui2021ffnet} & 3D       & Deep Learning & 87.28    \\
Zhu \textit{et al.} \cite{zhu2019discriminative} & 3D       & Deep Learning & 87.69    \\ \hline
Ours (Baseline)                & 3D       & Deep Learning & 84.83    \\
Ours                          & 3D       & Deep Learning & 89.15 $\pmb{(\uparrow 4.32)}$    \\ \hline
\end{tabular}
\end{table}

\begin{table}[htb]
\centering
\caption{Quantitative comparison on Bosphorus. \dag indicates that the results achieved by using only 3D modality are also reported.}
\label{table:Bos}
\begin{tabular}{cccc}
\hline
Method                        & Modality & Feature       & Accuracy(\%) \\ \hline
Tian \textit{et al.} \cite{tian20193d} & 2D+3D    & Deep Learning & 79.17    \\
Li \textit{et al.} \cite{li2015efficient} & 2D+3D    & Hand-crafted  & 79.72    \\
Li \textit{et al.} \cite{li2017multimodal} & 2D+3D    & Deep Learning & 80.28    \\
Wei \textit{et al.} \dag \cite{wei2018unsupervised} & 2D+3D    & Deep Learning & 82.50    \\
Jiao \textit{et al.} \cite{jiao20202d+} & 2D+3D    & Deep Learning & 83.63    \\
Ni \textit{et al.} \dag \cite{ni2022facial} & 2D+3D    & Deep Learning & 85.16    \\
Sui \textit{et al.} \dag \cite{sui2021ffnet} & 2D+3D    & Deep Learning & 87.65    \\
Sui \textit{et al.} \dag \cite{sui2023afnet} & 2D+3D    & Deep Learning & 88.31    \\
Lin \textit{et al.} \dag \cite{lin2020orthogonalization} & 2D+3D    & Deep Learning & 89.28    \\
Zhu \textit{et al.} \dag \cite{zhu2022cmanet} & 2D+3D    & Deep Learning & 89.36    \\ \hline
Wei  \textit{et al.} \cite{wei2018unsupervised} & 3D       & Deep Learning & 65.00    \\
Li   \textit{et al.} \cite{li20123d} & 3D       & Hand-crafted  & 75.83    \\
Yang \textit{et al.} \cite{yang2015automatic} & 3D       & Hand-crafted  & 77.50    \\
Ni   \textit{et al.} \cite{ni2022facial} & 3D       & Deep Learning & 77.82    \\
Zhu  \textit{et al.} \cite{zhu2022cmanet} & 3D       & Deep Learning & 81.25    \\
Sui  \textit{et al.} \cite{sui2023afnet} & 3D       & Deep Learning & 82.06    \\
Sui  \textit{et al.} \cite{sui2021ffnet} & 3D       & Deep Learning & 82.86    \\
Lin  \textit{et al.} \cite{lin2020orthogonalization} & 3D       & Deep Learning & 83.55    \\ \hline
Ours (Baseline)                & 3D       & Deep Learning & 83.23    \\
Ours                          & 3D       & Deep Learning & 86.77 $\pmb{(\uparrow 3.54)}$    \\ \hline
\end{tabular}
\end{table}

\subsection{Quantitative comparison}

\subsubsection{Results on BU-3DFE}

TABLE \ref{table:BU3D} provides a comparison  of our approach against state-of-the-art methods achieved on the BU-3DFE dataset, where the baseline counterpart entails the direct utilization of the expression branch trained in stage \romannumeral1, amalgamated with the classifier head for expression recognition. Remarkably, our method achieves an accuracy of $89.15\%$, surpassing other 3D-based approaches that utilize either hand-crafted or deep learning features. Furthermore, our method, which exclusively employs 3D data, closely approaches the performance achieved by 2D+3D multi-modal methods. To ensure an equitable evaluation, we also consider the results reported by these approaches when exclusively utilizing 3D data. The comparative findings unequivocally establish our method as the top performer in the 3D data-only scenario.

%Attention is drawn to the fact that our proposed method significantly improves the baseline method by $4.33\%$, which strongly indicates the effectiveness of disentangled learning.

\subsubsection{Results on Bosphorus}

TABLE \ref{table:Bos} presents the comparative results on the Bosphorus dataset. 
In a similar fashion, our approach notably surpasses other methodologies that exclusively rely on 3D data, underscoring its superior performance.
It is worth noting that while the work by Jiao \textit{et al.} \cite{jiao20202d+} merely provides experimental results employing multi-modal data, our method outperforms their approach by a large margin of $3.14\%$ on the Bosphorus dataset. 

\subsubsection{Comparison with the Baseline}

It is remarkable that our method demonstrates significant improvements of $4.32\%$ and $3.54\%$ when compared to the baseline models on the BU-3DFE and Bosphorus datasets, respectively. These results clearly illustrate that disentangled learning significantly improves the performance of our approach.

\begin{table*}[]
\caption{Ablation experiments of the loss functions, the fusion module, and the training strategy.}
\label{table:ablation}
\centering
\resizebox{0.8\textwidth}{18mm}{
\begin{tabular}{cccc|cc|ccc|c}
\hline
\multicolumn{4}{c|}{\textbf{Loss Function}}                             & \multicolumn{2}{c|}{\textbf{Network Structure}} &\multicolumn{3}{c|}{\textbf{Training Strategy}}      & \multirow{2}{*}{\textbf{Accuracy}} \\ \cline{1-9}
\textbf{w/o $L_{tri}$} & \textbf{w/o $L_{cls}$} & \textbf{w/ $L_{KL}$} & \textbf{w/ $L_{JS}$} & \textbf{w/o Fusion} & \textbf{w/o Skip Connection} & \textbf{Stage \romannumeral1} & \textbf{Stage \romannumeral2} & \textbf{Stage \romannumeral3}                                  \\ \hline
\checkmark        &                  &                 &                &                  &                 & \checkmark      & \checkmark     & \checkmark             &   85.92                            \\
                  & \checkmark       &                 &                &                  &                 & \checkmark      & \checkmark     & \checkmark             &   87.59                            \\
                  &                  & \checkmark      &                &                  &                 & \checkmark      & \checkmark     & \checkmark             &   85.98                            \\
                  &                  &                 & \checkmark     &                  &                 & \checkmark      & \checkmark     & \checkmark             &   85.04                            \\ \hline
                  &                  &                 &                & \checkmark       &                 & \checkmark      & \checkmark     & \checkmark             &    87.15                           \\ 
                  &                  &                 &                &                  & \checkmark      & \checkmark      & \checkmark     & \checkmark             &    87.76                           \\ \hline
                  &                  &                 &                &                  &                 & \checkmark      &                &                        &    84.83                           \\
                  &                  &                 &                &                  &                 & \checkmark      & \checkmark     &                        &    86.67                           \\
                  &                  &                 &                &                  &                 & \checkmark      & \checkmark     & \checkmark             &    \textbf{89.15}      \\ \hline                    
\end{tabular}}
\end{table*}

\subsection{Ablation Studies}

To further prove the validity of every component of our method, we conduct extensive ablation experiments on BU-3DFE, and the results are shown in TABLE \ref{table:ablation}.

\textit{Evaluation of the loss functions.}
In the latter two training stages, we employ the cross-entropy loss \eqref{eq_ce} and the triplet loss \eqref{eq1} to constrain the learning of the expression feature space. 
We remove these two parts separately for comparison and find that the classification accuracy decreases, pointing to the value of these two kinds of losses for the overall supervision.
Furthermore, drawing from prior research in the domain of disentanglement-based 3D face reconstruction \cite{olivier2023facetunegan}\cite{jiang2019disentangled}, we attempt to employ Kullback-Leibler (KL) divergence loss and Jensen-Shannon (JS) divergence loss as constraints to govern the feature space distribution. However, these endeavors do not yield favorable outcomes. The experiments suggest that the features extracted from the given point cloud data do not conform to the standard distribution of 3DMM, leading us to abstain from the utilization of KL loss and JS loss in our final framework configuration.

\textit{Evaluation of the network structure.}
We perform ablation experiments on the network structure to validate the design. 
Attention is given to the fusion module, which is a separate componet in the whole framework. It is used to supervise the completeness of the features learned by the two branches and we ablate this module separately to justify the design.
The experimental results reveal a reduction in classification accuracy upon its removal. 
The observation strongly indicates that the module plays a pivotal role in effectively encouraging the expression branch to acquire a more comprehensive understanding of expression-related information.
Further, we ablate the skip connection setting in the fusion module. 
The result shows that the fusion module lacking skip connections exhibits an improvement in performance compared to the baseline. 
Nevertheless, the degree of improvement is relatively lower compared to the fusion module incorporating skip connections. 
This observation implies that the inclusion of skip connections enables the integration of multi-scale information, thereby facilitating the fusion reconstruction task and ultimately enhancing the overall expression recognition capability of the framework.

\textit{Evaluation of the training strategy.}
We employ a multi-stage training strategy that allows the framework to learn disentangled expression feature space step by step. 
We test the classification accuracy of the features extracted by the expression branch at the end of each training phase and report the results. 
The incremental enhancements in classification accuracy achieved through the fine-tuning process serve as compelling evidence that each stage of our fine-tuning is effective and supports the rationale behind employing a multi-stage training strategy.

\begin{figure}[htb]
\centering
\includegraphics[width=0.48\textwidth]{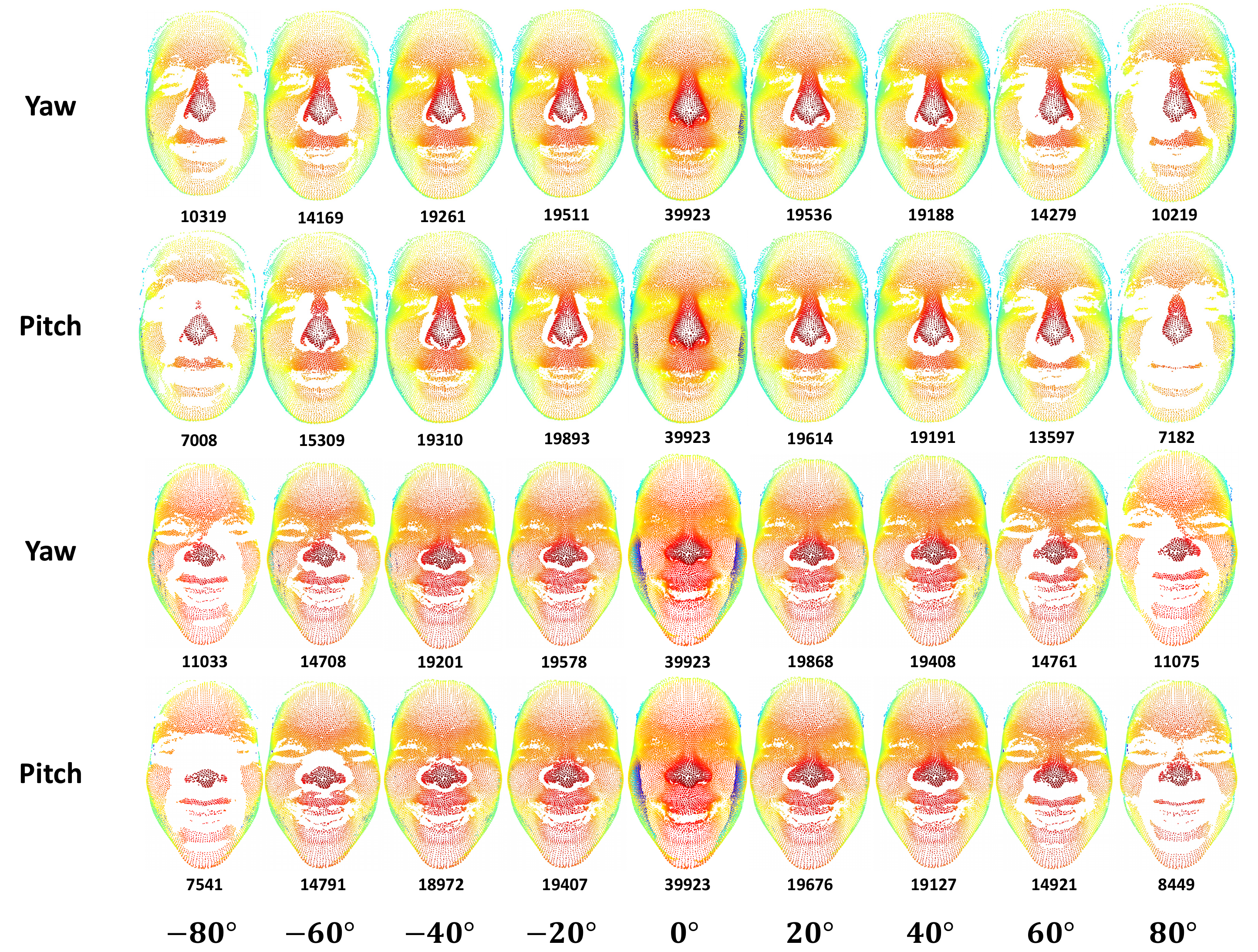}
\caption{Visualization of the rotated faces. The faces in the top two rows and the bottom two rows are from two different randomly selected subjects. Columns indicate different rotation angles. Faces in the same column have the same rotation angle. The numbers below each facial scan represent the points that are preserved after point cloud rotation.}
\label{fig:rotation-sample}
\end{figure}

\begin{figure}[htb]
\centering
\includegraphics[width=0.5\textwidth]{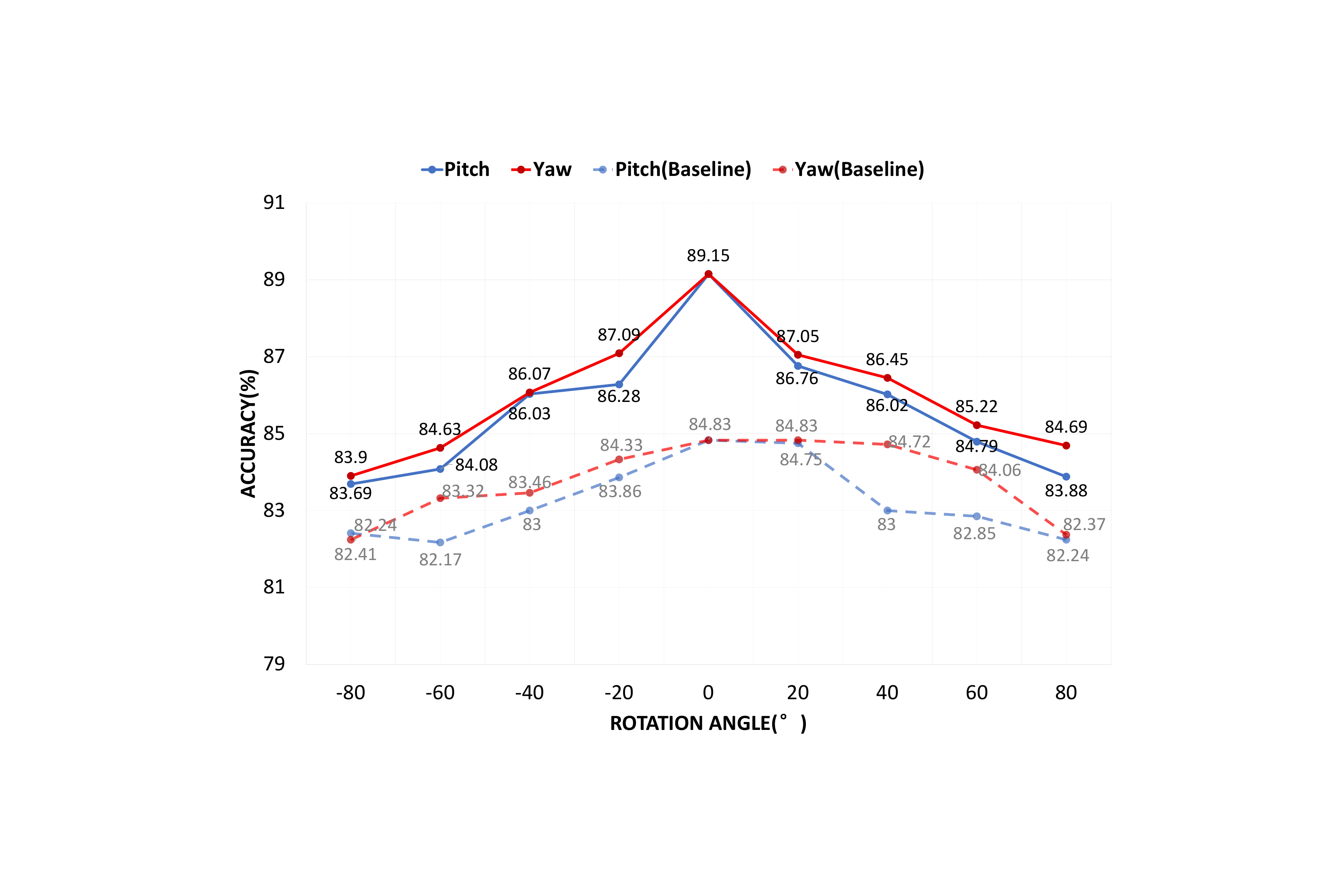}
\caption{Comparative results of the rotation experiment between the disentangled and the baseline methods. The results of these two methods are plotted as solid and dashed lines, where blue and red colors represent the pitch and yaw rotations, respectively. }
\label{fig:rotation-results}
\end{figure}

\subsection{Robustness to Rotation}
%Since only 3D data is used, our method is robust to facial pose changes, for which we carry out additional experiments on rotation data.
Our method exhibits robustness to facial pose changes, which can be attributed to the utilization of 3D mode data. To further validate this aspect, we conducted additional experiments on the rotated data. Given that the BU-3DFE dataset only consists of frontal faces, we artificially introduce variations by rotating the samples and subsequently removing invisible points to simulate self-occlusion.

We follow the general  experimental protocol on BU-3DFE and randomly select 60 subjects for a 10-fold cross-over experiment.
For each chosen subject, we rotate its 12 3D facial scans (six expressions, two highest intensities) to generate new data. 
We ultimately pick 16 rotations (pitch and yaw: $-80^{\circ}, -60^{\circ}, -40^{\circ}, -20^{\circ}, 20^{\circ}, 40^{\circ}, 60^{\circ}, 80^{\circ}$) and randomly select two samples for the presentation of rotated data as Figure \ref{fig:rotation-sample} shows.
As the rotation angle increases, the number of the occluded points increases and at the extreme angle ($\pm80^{\circ}$), significantly fewer points are remained in the pitch angle transformation than in the yaw angle transformation.

Figure \ref{fig:rotation-results} sequentially illustrates the recognition results for the 17 poses (plus $0^{\circ}$) that we test. 
Given that both the disentangled method and the baseline method utilize point cloud data as input, it is evident that their results demonstrate certain similarities. 
However, through a comprehensive evaluation, it becomes apparent that the disentangled approach exhibits a notable superiority over the baseline.
With increasing rotation angles, our recognition accuracy decreases, but all of them is maintained at a high level (above $80\%$) , which confirms the rotation-invariant capability of our method. 
We also notice that the variation of yaw angle has a smaller impact on the recognition accuracy, which we analyze to the fact that the left and right faces have some symmetry, and the network automatically focuses on the unoccluded other side in case of occluding one side.
In general, the results of the rotation experiment suggest that our method has promising potential for handling 3D FER in more practical scenarios.

\begin{figure}[htb]
\centering
\includegraphics[width=0.48\textwidth]{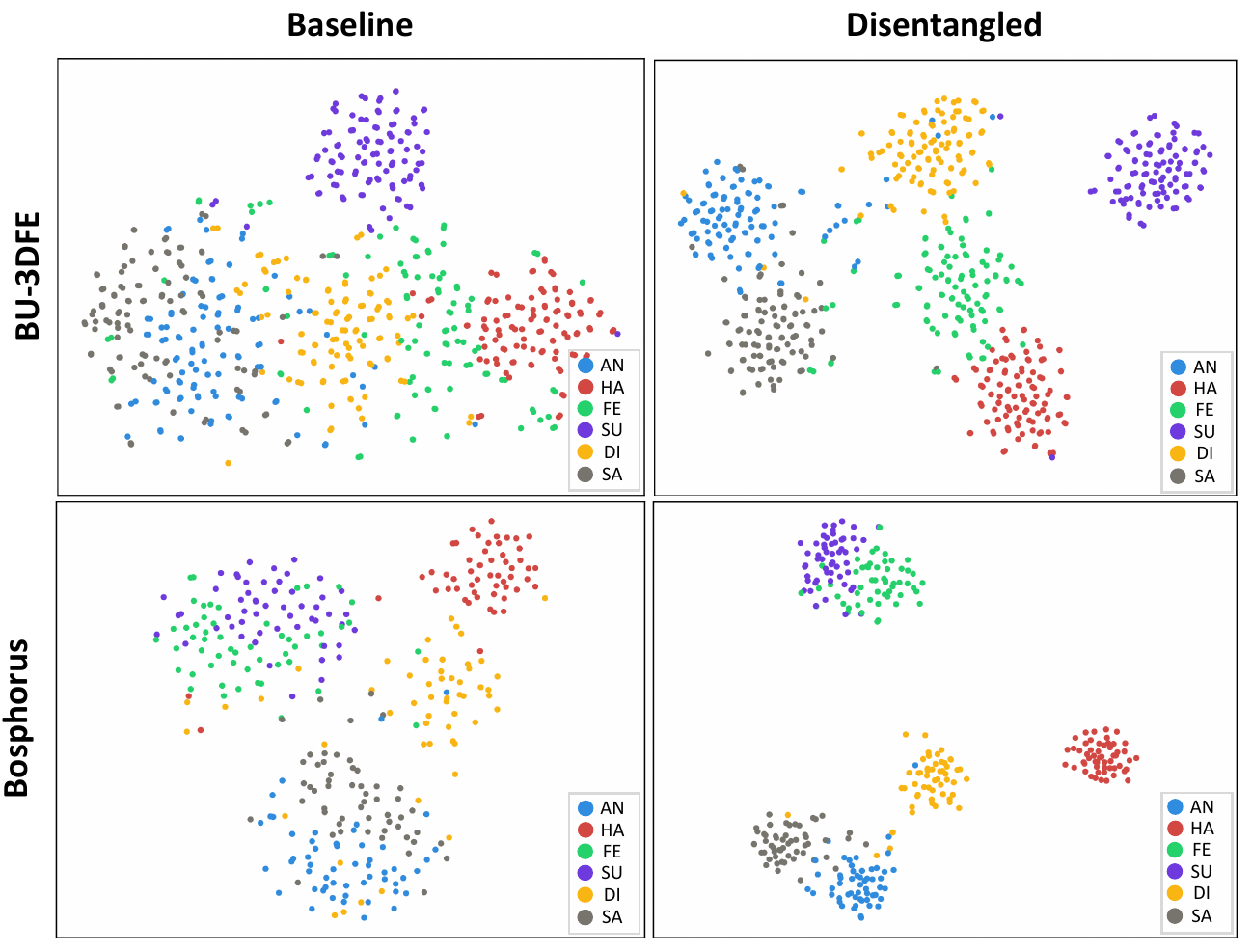}
\caption{T-SNE visualization of the expression features extracted by the baseline method and the proposed one on BU-3DFE and Bosphorus datasets. Six basic expressions are marked with different colors.}
\label{fig:t-sne}
\end{figure}

\subsection{Visualization}
%As a demonstration of the effectiveness of our disentanglement method, we visualize the expression features extracted by the baseline and the proposed method using t-SNE\cite{van2008visualizing}. 
%The first and second rows of Figure \ref{fig:t-sne} respectively correspond to the BU-3DFE and Bosphorus datasets, where the left column is before disentangling and the right column is after disentanglement. 
To exemplify the efficacy of our disentanglement method, we employ t-SNE \cite{van2008visualizing} to visualize the expression features extracted by both the baseline and our proposed method. In Figure \ref{fig:t-sne}, the first and second rows correspond to the BU-3DFE and Bosphorus datasets, respectively. The left column represents the features before disentanglement, while the right column illustrates the features after disentanglement.
It is readily apparent that the features extracted by the baseline method exhibit a lack of differentiation, characterized by fuzzy boundaries between expression categories. In contrast, our method yields features with dense clustering, demarcating clear and distinct boundaries for each expression category. These visualizations serve as compelling evidence that the disentangled expression features acquired through our approach are more amenable to classification tasks.
%One can clearly observe that the features extracted by the baseline method are poorly differentiated with fuzzy boundaries between categories, while the features extracted by ours have dense clustering and each expression category has distinct boundaries.
%The visualization results provide strong evidence that the disentangled expression features learned by our method are more easily classified.

\section{CONCLUSION}

%In this paper, we present a 3D FER method, DrFER, based on disentangled representation learning, which is practiced through a dual-branch framework suitable for point cloud data. 
%The cross-over structure and the fusion module in this framework drive the dual-branch network to learn expression and identity information, respectively. 
%In this study, we introduce the concept of disentangling to the field of 3D Facial Expression Recognition (FER) through our novel approach, DrFER. 
%DrFER is implemented using a dual-branch framework designed for 3D point cloud data. 
%Within this framework, the cross-over structure and fusion module facilitate the concurrent learning of expression and identity information by the two branches.
%In addition, the triplet loss and the cross-entropy loss adopted to further constrain the feature space for point clouds learned by the network. 
%Experimental results show that our method exceeds other state-of-the-art 3D FER methods and achieves close results to 2D+3D FER methods when only 3D point cloud data is utilized, all of which illustrate the effectiveness of the disentangling method.
%Our method addresses 3D FER through the process of disentangled representations learning, and its robustness to pose variations reveals a strong potential to tackle face expression recognition in realistic scenarios, as well as providing new insights for future work in 3D FER and 2D+3D FER.
In this study, we introduce the concept of disentangled representation learning into 3D FER via our approach, DrFER. 
The method utilizes a dual-branch framework for 3D point cloud data, where the cross-over structure and fusion module enable concurrent learning of expression and identity information. 
Additionally, triplet loss and cross-entropy loss constrain the feature space for improved results on 3D point clouds. 
Experimental results demonstrate its superiority over other 3D FER methods, achieving comparable results to 2D+3D FER methods when using only 3D data. 
DrFER addresses 3D FER through disentangled representations learning, demonstrating robustness to pose variations and providing insights for future research in 3D and 2D+3D FER.

\section{ACKNOWLEDGMENTS}

This work is partly supported by the National Key R\&D Program of China (No. 2022ZD0161902), the National Natural Science Foundation of China (No. 62176012, 62202031), the Beijing Natural Science Foundation (No. 4222049), and the Fundamental Research Funds for the Central Universities.

{\small
\bibliographystyle{ieee}
\bibliography{egbib}
}

\end{document}